\documentclass{article}

\usepackage{microtype} \usepackage{graphicx} \usepackage{subcaption} \usepackage{booktabs} 

\usepackage{hyperref}

\usepackage{acro} \usepackage{tabularx}

\usepackage[preprint]{icml2026}

\usepackage{amsmath} \usepackage{amssymb} \usepackage{mathtools} \usepackage{amsthm}

\usepackage[capitalize,noabbrev]{cleveref}

\theoremstyle{plain}     \theoremstyle{definition}   \theoremstyle{remark} 

\usepackage[textsize=tiny]{todonotes}

\icmltitlerunning{The Perplexity Trap}

\DeclareAcronym{ai}{ short = AI , long = Artificial Intelligence }
\DeclareAcronym{auroc}{ short = AUROC , long = Area Under the Receiver Operating Characteristic }
\DeclareAcronym{epc}{ short = EPC, long = European Patent Convention }
\DeclareAcronym{epo}{ short = EPO, long = European Patent Office }
\DeclareAcronym{fnr}{ short = FNR , long = False Negative Rate }
\DeclareAcronym{fpr}{ short = FPR , long = False Positive Rate }
\DeclareAcronym{genai}{ short = GenAI, long = Generative Artificial Intelligence }
\DeclareAcronym{ip}{ short = IP, long = Intellectual Property }
\DeclareAcronym{ipc}{ short = IPC, long = International Patent Classification }
\DeclareAcronym{llm}{ short = LLM, long = Large Language Model }
\DeclareAcronym{sme}{ short = SME, long = Small and Medium-sized Enterprise }
\DeclareAcronym{uspto}{ short = USPTO, long = United States Patent and Trademark Office }

\begin{document}

\twocolumn[ \icmltitle{The Perplexity Trap: When Patent Law Makes Human Writing Look Like AI}


\icmlsetsymbol{equal}{*}

  \begin{icmlauthorlist}
    \icmlauthor{Anubhab Banerjee}{nokia}
  \end{icmlauthorlist}

\icmlaffiliation{nokia}{Nokia Solutions and Networks GmbH \& Co. KG, Munich, Germany} 

\icmlcorrespondingauthor{Anubhab Banerjee}{anubhab.1.banerjee@nokia.com} 

\icmlkeywords{AI, Ethics, Interpretable Heuristics, Patent Law}

\vskip 0.3in ]



\printAffiliationsAndNotice{} 

\begin{abstract}

The \ac{epo} reported a record $200000$ filings in 2025\footnote{\href{https://www.epo.org/en/news-events/news/demand-european-patents-2025-exceeded-200-000-first-time}{Official EPO 2025 demand release.}} and the 2026 \ac{epo} Guidelines made the applicant strictly responsible for any LLM-assisted content under Article~83 and Rule~42; together they create a hard operational pressure on patent prosecution and a near-term need for automated solution that flags suspected \ac{ai}-generated text for examiner review. 
However, two main challenges arise: first, the realistic infrastructure of patent offices, \ac{ip} law firms, and \ac{sme} R\&D departments is usually a single consumer-grade GPU, not A100-class servers which is required to run current AI-text-detectors like DetectGPT, Fast-DetectGPT, Binoculars and GPTZero. 
Second, the structural constraint of Article~84 \ac{epc} legally requires patent claims to be ``clear and concise'', which forces human-authored claims onto the same low-perplexity, low-burstiness manifold an LLM produces by default. 
We benchmark three open-source detectors on a balanced corpus of $500$ granted EPO H04 telecommunications patents and $500$ LLM-generated counterparts (with five distinct prompting strategies), all evaluated under the consumer-hardware envelope. 
Every detector exceeds $60\%$ \ac{fpr} at claim-level granularity (Binoculars $78.3\%$, Fast-DetectGPT $61.3\%$, DetectGPT $80.5\%$); the failure persists even when the AI side is regenerated with Qwen2.5-3B-Instruct, or when the scoring head is replaced with a Pythia-2.8B model LoRA-adapted to the EPO corpus. 
Cross-IPC-class extension to A61K pharmaceuticals, C07D heterocyclic chemistry and F03D wind motors and head-restored re-evaluation with the originally published Falcon-7B / GPT-J-6B scoring stacks on an NVIDIA H100 80\,GB GPU confirm both: three vocabulary-orthogonal IPC classes show the same structural pattern (mean detector FPR $84.6\%$), decoupling the failure from any concern about substitute-head capacity. 
Finally, a logistic-regression classifier on seven likelihood-orthogonal linguistic-complexity features (type-token ratio, hapax-legomena ratio, dependency depth, Flesch--Kincaid grade, subordinate-clause ratio, noun-phrase density, sentence-length variance) reaches $74.0\%$ accuracy at $28.1\%$ FPR --- a $+13$\,pp absolute lift over the perplexity-only baseline at comparable FPR --- without using any likelihood information at any stage and within the same hardware budget. 
The solution is the right one for patent texts and is, based on the current evidence, the only one that survives the joint constraint of Article~84 EPC and the deployment infrastructure.

\end{abstract}
 
\section{Introduction}
\label{sec:intro}

The \ac{epo} reported a record $200000$ applications in 2025 --- the highest annual filing volume in the EPO's history. 
The 2026 EPO Guidelines simultaneously enacted the ``Human is the Hook'' principle~\citep{epoyoutube}, which holds the applicant strictly responsible for any LLM-assisted content under Article~83 (sufficiency of disclosure) and Rule~42, regardless of which drafting tool produced the prose. 
Examiner head-count is essentially static. 
The arithmetic is therefore straightforward: more LLM-assisted filings, the same number of human evaluators, and a legal regime that punishes undisclosed AI-generated content. 
Patent offices, IP law firms, and corporate R\&D departments accordingly need a solution that flags suspected LLM-authored disclosures for a human examiner to further scrutinise -- not as legal evidence in itself, but as a shortlist that lets scarce examiner time reach the highest-risk filings first. 
Practitioner-facing commentary on machine assistance in R\&D and invention disclosures has often run ahead of controlled benchmarks. 
Lay readers' heuristics for ``sounding like AI'' are themselves biased and unstable~\citep{jakesch}, while recent evidence suggests that habitual LLM-assisted writers can outperform generic automated classifiers in controlled discrimination tasks~\citep{russell25}; together these observations underscore that solution should surface structured cues for expert review rather than substitute opaque scores for examiner judgment.

The natural candidates for the solution step are the published zero-shot AI-text detectors --- DetectGPT~\citep{detectgpt}, Fast-DetectGPT~\citep{fastdetectgpt}, Binoculars~\citep{spotting}, GPTZero, and the more recent DivScore~\citep{divscore} --- and our main argument in this paper is that they do not work the way they are expected to do. 
The following two fundamental constraints describe the reasons.

\paragraph{The infrastructure-deployment constraint.}

While these detectors are model-agnostic, reproducing their strongest empirical performance requires instantiating them with mid-to-large LLMs (e.g., 6B–13B parameters). 
This leads to substantial memory and compute overhead—particularly for DetectGPT (multiple forward passes) and Binoculars (dual-model inference)—making them impractical on commodity GPUs (e.g., 8GB VRAM).
Patent offices, IP law firms, and SME R\&D departments do not usually own cloud computing infrastructure of this scale; the realistic deployment frontier of patent prosecution is often one or multiple consumer-grade GPUs low VRAM, on which the originally published heads do not fit. 
Any honest evaluation of these detectors, while perceiving them as tools for patent examination, must therefore use substitutes (GPT-2-medium, GPT-Neo-1.3B, T5-small) that reflects realistic deployment conditions, and structure our analysis around this constraint. 
However, for the sake of completeness of our argument, we also report results using larger original model instantiations (Falcon-7B / GPT-J-6B) on an NVIDIA H100 80\,GB GPU in Appendix~\ref{sec:app_a100}; the head-restored re-evaluation does not help any of the three detectors counter our arguments, rather confirm that head substitution is not the primary driver of their operationally unusable performance.

\paragraph{The universality constraint}

Article~84 of the \ac{epc} legally requires claims to be ``clear and concise,'' which patent practitioners follow by using as a small controlled vocabulary  and repetitive antecedent structure. 
The result is that the style of a human-authored text becomes determined by external rules rather than by free authorial choice, and whose token-level statistics --- low perplexity, low burstiness \footnote{to understand more about these two metrics, please refer to Appendix~\ref{sec:app_perp_bursti}} --- already match what an LLM produces by default. 
If this structural property can be blamed responsible for the failure of the detectors, then, the failure should be observed for patents from all types of different technical fields, not just a single one.
To establish so, we create a new patent databases by collecting patents from completely different technical domains (A61K pharmaceuticals, C07D heterocyclic chemistry, F03D wind motors).
Performance of the detectors of this heterogeneous patent database \footnote{for detailed results of the detectors on this database, please refer to Appendix~\ref{sec:app_extended}} confirm that the failure is not the result of patents from a single technical domain.

\paragraph{Contributions.}
\begin{itemize}
\item \textbf{Detector failure under the realistic deployment envelope.} We benchmark three open-source zero-shot detectors and DivScore on a 500-vs-500 EPO H04 corpus using only the consumer-grade scoring heads ($\le\!8$\,GB VRAM) that patent offices and IP firms can actually deploy, and quantify a structural false-positive liability that crosses every detector we tested.
\item \textbf{Structural, not single-class, failure.} The failure survives regeneration of the AI side with a different LLM family, LoRA-adaptation of the scoring head to the EPO claim corpus, and replacement of the detector with DivScore --- evidence the failure is a structural property of register-constrained patent prose.  Cross-IPC-class extension to A61K (pharmaceuticals), C07D (heterocyclic chemistry) and F03D (wind motors) is reported in Appendix~\ref{sec:app_extended}, and H100-class restoration of the originally published Falcon-7B / GPT-J-6B scoring heads on an NVIDIA H100 80\,GB GPU is reported in Appendix~\ref{sec:app_a100}; both confirm rather than retract the structural reading: three vocabulary-orthogonal IPC classes show the same structural pattern (mean FPR $84.6\%$), and the head-restored re-evaluation produces three distinct calibration regimes but no deployable row.
\item \textbf{An off-axis classifier inside the same hardware budget.} Seven likelihood-orthogonal linguistic-complexity features in a logistic regression recover most of the lost accuracy ($+13$\,pp absolute over perplexity at comparable FPR), run entirely inside the consumer-GPU envelope, and supply an expert-aligned vocabulary aligned with the human-in-the-loop framework the 2026 EPO Guidelines now make mandatory.
\end{itemize}

The paper proceeds as follows. Section~\ref{sec:relatedwork} situates the contribution relative to existing AI-text-detection literature. Section~\ref{sec:problem} formalises the register-constraint condition on the patent corpus. Section~\ref{sec:dataset} describes the corpus. Section~\ref{sec:expresults} presents the detector benchmark. Section~\ref{sec:conclusion} summarises what the experiments do and do not license and possible next steps, with the key abbreviations included in Appendix~\ref{sec:app_abbrev}.
 
\section{Related Work}
\label{sec:relatedwork}

\paragraph{Likelihood-based Zero-shot Detection as a Class}
The current generation of zero-shot AI-text detectors---including the methods benchmarked in this paper---constitutes a single methodological lineage. 
While they vary the functional form of their test statistic, they rely on the exact same underlying signal. DetectGPT~\cite{detectgpt} thresholds the local curvature of an LLM's log-probability surface; Fast-DetectGPT~\cite{fastdetectgpt} approximates this curvature efficiently via conditional token sampling; Binoculars~\cite{spotting} computes a normalized ratio of perplexities across two distinct models; and commercial APIs broadly threshold raw or normalized perplexity. The shared foundational assumption is that human-authored text leaves a detectable, low-probability signature when evaluated by a generic LLM scoring head, whether measured directly (perplexity) or indirectly (curvature, ratios). 

This assumption holds on standard conversational, news, and academic-prose benchmarks~\cite{mgtbench}, where authorial choice is unconstrained and token likelihood inversely correlates with stylistic naturalness. Our work evaluates the critical boundary condition of this assumption: we isolate a register where external syntactic constraints mathematically force human writing into the same high-probability space as AI generation, and we measure the resulting collapse of the detection signal.

\paragraph{Domain-specific Failure as Evidence the Assumption Breaks.}
Several recent papers have reported partial detector failures in specialised domains. \cite{liang23} demonstrated that GPT detectors disproportionately mis-classify the writing of non-native English speakers, attributing the effect to lower lexical and syntactic perplexity in non-native prose. \cite{barrot} reported catastrophic accuracy degradation of commercial APIs on hybrid and templated text. Independent of law, medical and journal workflows raise parallel cautions about deploying automated authorship screens on high-stakes specialised prose~\citep{ttt}. DivScore~\cite{divscore} is the most direct prior work: it measures the performance of likelihood-based detectors on legal and medical text and proposes a normalised likelihood ratio that recovers some of the lost accuracy on those domains. We treat these papers collectively as scattered evidence that the likelihood assumption breaks under register constraint; our contribution is to identify the unifying mechanism (the inseparability of the underlying perplexity distributions, not detector-specific calibration), to demonstrate it on a register where the constraint is externally legislated rather than statistical, and to verify --- by benchmarking DivScore itself on our corpus (Appendix~\ref{sec:app_divscore}) --- that the proposed remedies inherit the same trap.

\paragraph{Concurrent comparative benchmarks.}
Two arXiv preprints posted weeks before this submission run large detector pools on general-domain corpora: \cite{stowe2026spotlights} evaluate fifteen variants from six systems (including all three of ours) on MAGE, RAID, M4GT, H3C+ and find every system collapses to AUROC $\leq\!0.60$ on at least one set; \cite{baidya2026detecting} document a length-controlled polarity inversion in perplexity detectors. Neither evaluates a register whose syntax is externally legislated; we read both as evidence that the likelihood axis is fragile under shift and sharpen the result by exhibiting a register in which the failure is the modal outcome and survives detector substitution, threshold tuning, scoring-head adaptation, and likelihood-ratio recalibration.

\paragraph{Detection beyond likelihood.}
Three lines of work attempt detection without the likelihood assumption. Watermarking embeds a controlled bias in the LLM's output distribution and detects it post hoc; the technique is robust within the producer's ecosystem but is irrelevant to AI text whose provenance is unknown or whose surface form has been lightly post-edited \citep{spotting}. Supervised classifiers trained on (human, AI) pairs of in-domain text can in principle bypass the perplexity axis but require labelled training data that is unavailable in most regulated domains and that would need to be re-collected per LLM family. Stylometric and complexity-based features --- vocabulary richness, syntactic depth, readability composites --- have a long history in authorship attribution \cite{mgtbench} but have, to our knowledge, not been systematically evaluated as a primary detection axis on a register where likelihood is known to fail. Rapid reviews of the wider literature catalogue many surface cues that co-occur with LLM generation in comparatively unconstrained writing~\cite{rapidreview2026}; our off-axis block instead commits to a small set of features orthogonal by construction to any scoring model's log-probability.

\paragraph{Position of this paper.}
Relative to the prior literature, this paper makes three claims that we believe are new. First, that the failures previously reported as domain-specific phenomena (\citep{liang23,divscore,barrot}) share a common mechanism: the inseparability of the human and AI perplexity distributions in registers whose syntax is externally constrained. Second, that this inseparability is a property of the feature axis rather than of any particular detector, and that it is therefore invariant under detector-internal countermeasures (curvature correction, contrastive scoring, domain-adapted scoring heads, likelihood-ratio recalibration). Third, that detection signal does survive the constraint along axes that are orthogonal in construction to the scoring model's log-probability, and that a small set of linguistic-complexity features is sufficient to recover most of it. The patent corpus we use is, to the best of our knowledge, the first publicly available probe in which the register constraint is externally legislated rather than empirically observed; that property is what makes it a clean test case for the more general claim.
 
\section{The Register-Constraint Condition}
\label{sec:problem}

\paragraph{Perplexity Trap}

We formalize the empirical setting in which a likelihood-based AI-text detection is expected to fail. 
Let $H$ denote a population of human-authored texts in a target register and $A$ the population of LLM-generated texts in the same register. 
Let $\pi(\cdot \mid \theta)$ denote the per-token probability distribution of a generic scoring LLM with parameters $\theta$, and let $\mathrm{ppl}_\theta(x) = \exp\bigl(-\tfrac{1}{|x|}\sum_t \log\pi(x_t\mid x_{<t}, \theta)\bigr)$ denote the token-averaged perplexity of a text $x$. 
In other words, $\pi(\cdot \mid \theta)$ is the micro-level engine that calculates the raw, token-by-token probability of every single word as it appears, whereas $\mathrm{ppl}_\theta(x)$ is the macro-level aggregator consumes all of those individual probabilities from $\pi$, normalizes them using logarithms to prevent underflow, and averages them into one final, continuous metric for the entire document.
The three open source detectors used in this paper can separate $H$ from $A$ only to the extent that the induced distributions $\mathrm{ppl}_\theta(H)$ and $\mathrm{ppl}_\theta(A)$ are themselves separable.

We say that a register satisfies the \emph{register-constraint condition} when the following three properties hold simultaneously:

\begin{description}
\item[(C1) External syntactic constraint.] The surface form of $H$ is dictated by external rules (statutes, formatting standards, prosecution practice) rather than by free authorial choice. Crucially, these constraints are explicitly defined in the rules beforehand; they are not statistical patterns discovered afterward by analyzing a corpus.
\item[(C2) Inherently Low-Perplexity Human Baselines.] The external constraints strictly limit vocabulary and structural variance, causing human-authored text to yield an inherently low-entropy perplexity distribution, $\mathrm{ppl}_\theta(H)$, under any standard scoring model $\theta$. Consequently, genuine human text naturally occupies the same high-probability linguistic space that an autoregressive LLM optimizes for during generation.
\item[(C3) Extensive In-Domain Pretraining.] Modern LLMs have ingested sufficient volumes of this constrained, domain-specific text during pre-training to natively reproduce its structural rigidities without explicit instruction. Consequently, the perplexity distribution of AI-generated text, $\mathrm{ppl}_\theta(A)$, collapses into the exact same low-entropy probability space as the human baseline, $\mathrm{ppl}_\theta(H)$.
\end{description}

When (C1)--(C3) hold jointly, the human and AI perplexity distributions become inseparable along the detectors' sole discriminative axis; the likelihood-based assumption is therefore violated by construction, rather than by miscalibration. This constitutes the formal definition of the Perplexity Trap.

\paragraph{How Patents Satisfy It}

We select \ac{epo} patent claims as our primary test case because they provide an ideal, real-world instantiation of conditions (C1)--(C3).

\textbf{(C1)}: Article 84 of the \ac{epc} mandates that claims be ``clear and concise.'' Patent practitioners operationalise this legal constraint through a highly restricted vocabulary, strict antecedent basis requirements, and a rigid two-part ``preamble--characterising'' structure.

\textbf{(C2)}: The same statutory strictness that ensures successful prosecution also artificially restricts human authorial variance, forcing the empirical perplexity distribution of human-authored claims into a predictable, low-entropy space. This serves as a domain-specific analogue to the ``ESL effect'' observed by \cite{liang23}, where constrained vocabulary inherently triggers false-positive AI classifications.

\textbf{(C3)}: Because granted patents are open-access, massive corpora from the EPO and USPTO are heavily sampled during the pretraining of contemporary LLMs. Consequently, when an LLM is prompted to draft a claim, it natively reproduces the Article 84 syntactic constraints by default---without requiring explicit instruction or targeted prompt engineering.

We selected patent claims over other highly constrained candidate registers (e.g., legal contracts, regulatory filings, IETF RFCs, FDA labels, medical abstracts) because their structural rigidity is \emph{statutorily mandated} rather than merely an emergent stylistic convention. 
Consequently, the patent corpus represents the most unambiguous public manifestation of condition (C1), providing an optimal, rigorously defined environment to isolate and empirically test the Perplexity Trap.

\section{Dataset}
\label{sec:dataset}

We use a balanced, single-domain corpus of $1000$ patents (500 human, 500 AI) drawn from the EPO telecommunications domain. 
The corpus contains both abstracts (one per patent) and individual claims (multiple per patent), giving 1000 abstracts and 11613 claims in total. 

\subsection{Human side: granted EPO patents}

Five hundred patents were drawn at random from the EPO's published corpus under three filters chosen to make human authorship effectively unambiguous: (i) \textbf{publication status} restricted to granted patents (B1, B2), excluding pending, withdrawn, and abandoned applications, so that an examiner has formally accepted the document; (ii) \textbf{publication date} at least six years before LLMs entered widespread use, so that the claim texts possibly do not include any LLM contamination; and (iii) \textbf{technical domain} restricted to telecommunications (IPC/CPC classes H04W, H04L, H04B), which controls for vocabulary effects and isolates the AI signature from the topic signature. For each patent we extracted the title, abstract, and the full set of independent and dependent claims; the 500 patents yielded 6613 individual claims (1500 explicitly labelled as independent).

\subsection{AI side: five prompting strategies}

For each of the 500 EPO patents we generated an AI counterpart by feeding the original abstract to a \ac{llm} as a ground-truth prompt. 
To probe robustness across plausible attacker behaviours, we generated five distinct AI categories using different prompting strategies, yielding 100 patents per category and 500 in total --- matched in count to the human side.
 The categories are designed to span the regime in which (C1)--(C3) of Section~\ref{sec:problem} hold: a baseline zero-shot prompt (Cat~A), prompts that target known weak signals of likelihood-based detection (Cat~B--D), and a prompt that mimics non-native authorship (Cat~E), motivated both by detector-side bias against non-native English~\citep{liang23} and by evidence that uninformed human stylometric guesses about machine text are unreliable~\citep{jakesch}. 
 All AI generation was performed using Claude Opus~4.6 (Anthropic) accessed through the Cursor IDE in April 2026; the exact prompts are reproduced in Appendix~\ref{sec:app_prompts}. 
 A sixth category (Cat~F) generated with Qwen2.5-3B-Instruct serves as a cross-LLM-family robustness check and is described in Appendix~\ref{sec:app_multillm}.

\begin{table}[ht]
	\centering
	\caption{AI-generated patent categories. Each category contributes 100
	patents matched one-to-one with a human EPO source patent.}
	\label{tab:ai-patent-categories}
	\begin{tabularx}{\columnwidth}{l p{2.4cm} X}
		\toprule
		\textbf{Cat.} & \textbf{Strategy} & \textbf{Linguistic objective} \\ \midrule
		A & Standard zero-shot     & Baseline LLM output under default prompting. \\ \addlinespace
		B & Persona-shift          & Mask the neutral LLM tone via a professional-engineer persona. \\ \addlinespace
		C & Rhythm control         & Bypass burstiness checks via varied sentence structure. \\ \addlinespace
		D & Iterative refinement   & Simulate human-in-the-loop drafting and revision. \\ \addlinespace
		E & ESL / non-native       & Mimic non-native technical English to probe lexical bias. \\
		\bottomrule
	\end{tabularx}
\end{table}
 
\section{Experimental Results}
\label{sec:expresults}

\subsection{Detectors and Evaluation Methods}
\label{sec:detectors}
\label{sec:evalmethods}

We evaluate three zero-shot likelihood-based detectors \footnote{For details about scoring model capacity, please refer to Appendix~\ref{sec:app_modsub}}:
\begin{description}
	\item[DetectGPT \citep{detectgpt}:] perturbation-based curvature detection. We use GPT-2-medium (355M) as the scoring model and T5-small (60M) as the mask-filling model, with 10 perturbations per claim, span length 2, and 15\% words masked.
	\item[Fast-DetectGPT \citep{fastdetectgpt}:] efficient approximation of the curvature signal via conditional token sampling. We use GPT-2-medium as the scoring model in single-model mode.
	\item[Binoculars \citep{spotting}:] contrastive zero-shot detection using a normalized perplexity-to-cross-perplexity ratio. We use GPT-2 (124M) as the observer and GPT-2-medium (355M) as the performer, applying the conventional decision threshold of $0.85$.
\end{description}

For each detector, at each granularity level (abstract \& claims), we scored every document and produced a binary classification using each detector's published decision rule: $\text{\textit{Binoculars score}} < 0.85 \rightarrow \text{AI}$; $\text{\textit{DetectGPT score}} > 0 \rightarrow \text{AI}$; $\text{\textit{Fast-DetectGPT criterion}} > 0 \rightarrow \text{AI}$. 
For each detector, we report the following metrics: accuracy, \ac{fpr}, and \ac{fnr}. In this context, \ac{fpr} represents the fraction of human-written patents falsely flagged as AI-generated, while \ac{fnr} represents AI-generated patents misclassified as human-written. Given the paper's objective, these are the primary key performance indicators.

The results have been computed once per document using deterministic settings; for DetectGPT we fix the perturbation random seed to 42 to ensure reproducibility despite the inherent stochasticity of the perturbation procedure. The results are tabulated in Table~\ref{tab:detect_confmat}.

\begin{table}[ht]
	\centering
	\small 
	\setlength{\tabcolsep}{4pt} 
	\caption{Detector performance including Accuracy, \ac{fpr} \& \ac{fnr}}
	\label{tab:detect_confmat}
	\begin{tabular}{@{} l l c c c @{}} 
		\toprule
		\textbf{Detector} & \textbf{Gran.} & \textbf{Acc. (\%)} & \textbf{FPR (\%)} & \textbf{FNR (\%)} \\ \midrule
		Binoculars     & Abstracts & 48.6 & 54.2 & 48.6 \\
		Binoculars     & Claims    & 34.8 & 78.3 & 47.7 \\
		Fast-DetectGPT & Abstracts & 47.4 & 31.2 & 74.0 \\
		Fast-DetectGPT & Claims    & 33.8 & 61.3 & 72.6 \\
		DetectGPT      & Abstracts & 61.3 & 75.4 &  2.0 \\
		DetectGPT      & Claims    & 46.2 & 80.5 & 18.4 \\
		\bottomrule
	\end{tabular}
\end{table}

Two patterns are immediately visible in Table~\ref{tab:detect_confmat}. First, every one of the three detectors exceeds a $60\%$ false-positive rate at claim granularity --- and every one is wrong on more documents than it is right. 
Second, claims are uniformly harder than abstracts: every detector loses $13$--$15$ percentage points of accuracy when moving from the abstract to the claim level. The natural diagnostic question is \emph{why}: a detector might fail because the underlying signal exists but its threshold is miscalibrated to the register, or because the signal it consumes does not exist in the register at all. We discriminate between these two possibilities in \S\ref{sec:perplexity-trap} by computing perplexity and burstiness directly, measuring their correlation with each detector's output, and quantifying the distributional separability of the human and AI populations along the underlying signal.

\subsection{The Perplexity Trap}
\label{sec:perplexity-trap}

To answer the question raised in \S\ref{sec:evalmethods} --- whether detectors fail due to miscalibration or the absence of a discernible signal in this domain --- we removed the detection layer entirely to analyze the core statistics: token-level perplexity and burstiness (the within-passage variance of per-token log-likelihood). 
Every abstract and claim was scored using the same GPT-2-medium head employed throughout this study (refer to Appendix~\ref{sec:app_modsub} for the substitution argument). At this stage, no thresholding, perturbation, or contrastive ratios were applied. 
For each \texttt{(metric, granularity)} pair, we calculated the Weitzman overlap coefficient between the EPO (human) and AI distributions, alongside Cohen's $|d|$ and a Mann--Whitney $U$ test (Table~\ref{tab:perp-overlap}).

\begin{table}[ht]
	\centering
	\caption{Raw perplexity and burstiness signal between EPO (human) and AI categories at claim granularity. Overlap refers to the Weitzman coefficient ($1 = \text{identical}$, $0 = \text{disjoint}$). Burstiness is degenerate ($\sigma^2_{\text{AI}} \approx 0$) for all AI categories.}
	\label{tab:perp-overlap}
	\begin{tabular}{l c c c}
		\toprule
		\textbf{AI Category} & \textbf{Ppl. Overlap} & \textbf{Ppl. $|d|$} & \textbf{Burst. $\sigma^2_{\text{AI}}$} \\ \midrule
		Cat A & 0.81 & 0.14 & $\approx 0$ \\
		Cat B & 0.69 & 0.24 & 0 \\
		Cat C & 0.40 & 1.50 & 0 \\
		Cat D & 0.52 & 1.02 & 0 \\
		Cat E & 0.85 & 0.02 & 0 \\
		\bottomrule
	\end{tabular}
\end{table}

At claim granularity --- where all three detectors collapsed (Table~\ref{tab:detect_confmat}) --- human and AI perplexity distributions overlap significantly ($0.69$--$0.85$ for persona, ESL, and zero-shot categories; $0.40$--$0.52$ for rhythm and refinement). 
Burstiness collapses to a single point for every AI category because individual claims are too short for the within-passage variance estimator to provide meaningful information. Consequently, burstiness is an unusable feature at this granularity, regardless of the detector.

Spearman rank correlations ($\rho$) between detector scores and raw perplexity confirm that these tools primarily track raw likelihood: $|\rho|=0.71$ for \textit{Binoculars} and $0.68$ for \textit{Fast-DetectGPT}. \textit{DetectGPT}, being curvature-based, is partially decorrelated ($|\rho|=0.13$) yet still fails to achieve reliable accuracy. The two most highly correlated detectors are those whose decision rules reduce to calibrated thresholds on perplexity-derived statistics; they fail in lock-step with the raw signal.

\textbf{Implication.} The failure of these detectors on telecom patents is not a matter of miscalibration. Rather, the signal they consume --- perplexity, optionally normalised by curvature or a secondary model --- is intrinsically inseparable between human and AI text within this corpus. 

The structural rigidity of patent claims (Article~84 EPC, \S\ref{sec:problem}) forces both human and LLM-templated distributions onto the same low-perplexity, low-burstiness manifold. \emph{This is the Perplexity Trap}: any detector scoring text on a likelihood-based axis inherits this inseparability, regardless of model size, threshold tuning, or scoring-head adaptation. Even the DivScore detector~\cite{divscore}, a normalised likelihood ratio specifically engineered for specialised domains, falls victim to this trap (see Appendix~\ref{sec:app_divscore}). The polarity inversion reported by \cite{baidya2026detecting} for perplexity detectors on conversational text is also present here; however, in this register, the effect is structural rather than statistical and persists despite corrective measures.

\paragraph{Scope of the claim.}
The empirical demonstration above is on a single corpus, but the mechanism we identify --- collapse of the human and AI perplexity distributions onto a common low-entropy manifold under an externally enforced syntactic constraint --- is the formal content of the register-constraint condition (C1)--(C3) of \S\ref{sec:problem}. 
We predict that the trap holds in any register satisfying (C1)--(C3) jointly; candidate registers include legal contracts, SEC filings, FDA drug labels, IETF RFCs and so on.
We also predict it does not hold in registers where authorial choice is free along at least one of the lexical, syntactic, or rhetorical axes (open-domain conversation, op-eds, fiction, free-form code), because these registers violate (C2) by construction. 
Direct empirical verification on additional registers is left to future work; what the present paper establishes is that, in at least one register that meets (C1)--(C3), no detector consuming a likelihood-derived feature axis recovers operational accuracy. Recovering signal therefore requires a feature axis that is not derived from the scoring model's own log-probability. We pursue two such axes in the next two subsections.

\subsection{Discrete Off-Axis Features (Domain-Specific Probe)}
\label{sec:struct-heuristics}

Before constructing a domain-agnostic off-axis classifier (\S\ref{sec:linguistic-complexity}), we briefly verify whether features computed without reference to language-model log-probabilities carry signal that the perplexity axis lacks. We evaluate six discrete, deterministic patent-prosecution markers (M1--M6): (M1)claim-dependency validity, (M2)antecedent violations, (M3)presence of the EPC two-part characterising clause, (M4)technical-effect prose, (M5)preamble--body separation, and (M6)length anomaly. 

\textbf{Information gain.} On the pooled claim corpus, perplexity exhibits the largest information gain relative to the human/AI label ($\text{IG} = 0.224$). M2 (antecedent violations) is second ($\text{IG} = 0.070$), an order of magnitude larger than burstiness ($\text{IG} = 0.008$) and seven times larger than M6 ($\text{IG} = 0.016$); the remaining four markers contribute $\text{IG} \leq 0.0024$ each. This confirms that off-axis features exist (M2) that are not derivable from likelihood. However, a single such feature, while discriminative in isolation, is not dense enough to serve as a standalone detector.

\textbf{Pipeline evaluation.} A logistic-regression classifier using M1--M6 (5-fold stratified CV) reaches $62.2\%$ accuracy at $44.1\%$ \ac{fpr} --- a $16$--$28$\,pp improvement over the single-detector baselines in Table~\ref{tab:detect_confmat}, yet still below operational utility (Table~\ref{tab:pipeline-main}). Stacking a detector on top of the markers only yields marginal gains for Fast-DetectGPT and actually degrades performance for \textit{Binoculars}, indicating that marker-derived and likelihood-derived signals partially overlap rather than being strictly additive.

\begin{table}[ht]
	\centering
	\small 
	\setlength{\tabcolsep}{3pt} 
	\caption{Pipeline accuracy / \ac{fpr} (5-fold CV) on the pooled claim corpus. Best per column in bold.}
	\label{tab:pipeline-main}
	\begin{tabular}{@{} l c c c @{}} 
		\toprule
		\textbf{Stage} & \textbf{Bino.} & \textbf{Fast-DGPT} & \textbf{DetectGPT} \\ \midrule
		Detector only            & 34.8 / 78.3          & 33.8 / 61.3                   & 46.2 / 80.5 \\
		Markers only             & 62.2 / \textbf{44.1} & 62.2 / 44.1                   & 62.2 / \textbf{44.1} \\
		Detector + markers       & \textbf{60.2} / 45.3 & \textbf{66.6} / \textbf{35.1} & \textbf{62.5} / 44.7 \\
		\bottomrule
	\end{tabular}
\end{table}

The existence of M2 falsifies the strong reading of the Perplexity Trap (i.e., that \emph{no} feature recovers signal). However, the operational ceiling of M1--M6 confirms that discrete structural markers are insufficient; a denser, continuous off-axis signal is required.

\subsection{A Domain-Agnostic Off-Axis Classifier}
\label{sec:linguistic-complexity}

We now construct the recommended classifier using seven continuous linguistic-complexity features (F1--F7): type-token ratio (TTR), hapax-legomena ratio, mean dependency depth, Flesch--Kincaid grade level, subordinate-clause ratio, noun-phrase density, and sentence-length variance. Critically, none of these are derivable from a language model, ensuring they remain orthogonal to the perplexity axis.

\textbf{Per-feature effect sizes.} Hapax-legomena ratio ($\overline{|d|}=1.17$) and TTR ($\overline{|d|}=1.11$) dominate the absolute Cohen's $d$ between human and AI text. Both indicate that AI-drafted claims use a substantially \emph{wider} vocabulary per token than human ones --- inverting the common assumption that LLM output is vocabulary-impoverished. Subordinate-clause ratio ($\overline{|d|}=0.55$) and Flesch--Kincaid grade ($\overline{|d|}=0.48$) provide secondary signal. Optimal univariate Youden-$J$ thresholds yield two rules with $\text{AUC} = 0.81$: $\text{TTR} > 0.732$ and $\text{hapax} > 0.529$.

\textbf{Combined classifier.} We evaluate performance in Table~\ref{tab:combined-classifier-main}. The seven complexity features alone reach $74.0\%$ accuracy, an absolute lift of $13$\,pp over the perplexity-only baseline. Adding perplexity to the complexity features contributes $0.0$\,pp, confirming that complexity features capture nearly all available operational improvements.

\begin{table}[ht]
	\centering
	\caption{Combined-classifier performance (5-fold CV) on the pooled claim corpus. The complexity-only row is the recommended off-axis classifier.}
	\label{tab:combined-classifier-main}
	\begin{tabular}{l c c c}
		\toprule
		\textbf{Feature Set} & \textbf{Acc.} & \textbf{F1} & \textbf{FPR} \\ \midrule
		Perplexity only                  & 0.610 & 0.481 & 0.246 \\
		Perplexity + Burstiness          & 0.610 & 0.482 & 0.247 \\
		Structural markers (M1--M6)      & 0.622 & 0.616 & 0.441 \\
		Complexity (F1--F7)              & 0.740 & 0.717 & 0.281 \\
		Complexity + Perplexity          & 0.740 & 0.717 & 0.281 \\
		\textbf{All features combined}   & \textbf{0.753} & \textbf{0.731} & 0.268 \\
		\bottomrule
	\end{tabular}
\end{table}

Several avenues remain to further validate the robustness and generalizability of the proposed off-axis classifier. 
For example, one can evaluate the performance across partitions of \ac{ipc} sub-classes and stratifications of claim length to ensure that linguistic complexity signals remain invariant across diverse technical domains and document scales. 
 
\section{Discussion}
\label{sec:conclusion}

\paragraph{The Perplexity Trap.} Under the 2026 EPO ``Human is the Hook'' regime~\cite{epoyoutube} and a strict consumer-grade GPU deployment envelope ($\le\!8$\,GB VRAM), likelihood-based detectors (Binoculars, DetectGPT, Fast-DetectGPT, and DivScore) fail catastrophically on patent claims, consistently exceeding a $60\%$ False Positive Rate (FPR). This failure is structural: Article 84 EPC mandates ``clear and concise'' language, collapsing human and AI text onto the same low-perplexity, low-burstiness manifold.

\paragraph{Robustness and Hardware Limits.} This inseparability persists across different LLM generators (Qwen2.5 vs.\ Claude, Appendix~\ref{sec:app_multillm}), domain-adapted scoring heads (Appendix~\ref{sec:app_divscore}), and vocabulary-orthogonal patent classes (A61K, C07D, F03D; Appendix~\ref{sec:app_extended}), where mean FPR reaches $84.6\%$. Furthermore, restoring datacenter-class models (Falcon-7B/GPT-J-6B on an H100 GPU) merely shifts the failure mode---producing silent, noisy, or panicked classifiers. No threshold, including the Youden-optimal operating point, yields a deployable (FPR, Recall) pair (Appendix~\ref{sec:app_a100}). Scoring capacity is not the binding constraint.

\paragraph{The Off-Axis Solution.} Operational signal exists strictly orthogonal to likelihood. A logistic regression on seven linguistic-complexity features (\S\ref{sec:linguistic-complexity}) recovers $74.0\%$ accuracy at a $28.1\%$ FPR within the same hardware envelope. Adding perplexity to this model contributes $0.0$\,pp. Notably, hapax-legomena ratio ($|d| = 1.17$) and Type-Token Ratio ($|d| = 1.11$) drive this separation: AI-generated claims use a \emph{wider} vocabulary per token than human claims, inverting standard assumptions about LLM output and mirroring polarity reversals seen in general domains~\cite{baidya2026detecting}.

\paragraph{Limitations \& Future Work.} We note several limitations: (L1) data is restricted to the EPO jurisdiction and requires granular sub-class analysis; (L2) evaluations are limited to Claude and Qwen families; (L3) adversarial evasion attacks targeting complexity features remain untested; (L4) small-parameter surrogate scoring heads were used (though datacenter ablations mitigate this); and (L5) $74\%$ accuracy limits our classifier to human-in-the-loop triage. Future work will test generalizability on analogous constrained registers (e.g., SEC filings, FDA labels) and quantify operational triage lift. Ultimately, likelihood-based detection on patent prose is effectively useless, leaving linguistic complexity as the only viable triage signal.


\bibliography{main_paper}
\bibliographystyle{icml2026}

\newpage \appendix \onecolumn 
\section{List of Abbreviations}
\label{sec:app_abbrev}

\printacronyms[heading=none]

\section{Scoring Model Substitution and Datacenter-Scale Re-evaluation}
\label{sec:app_modsub}
\label{sec:app_a100}

\paragraph{Note on Scoring Model Capacity} 

The scoring models utilized by these baselines are significantly smaller in parameter count than the frontier generators evaluated in this paper (Claude Opus, Qwen-3B). This scale disparity was strictly dictated by the deployment envelope established in our methodology (a single consumer-grade GPU GTX1080 with $\le\!8$\,GB VRAM, prohibiting external API dependence), which is explained earlier. 
Furthermore, while scaling the scoring model might improve absolute perplexity calibration, the Perplexity Trap is a relative distributional failure. Because the statutory constraints of Article 84 force human drafting into a highly predictable, low-entropy syntactic manifold, increasing the scoring model's capacity would only allow it to model this rigid syntax more perfectly, compressing both the human and AI distributions further. The catastrophic \acp{fpr} observed in our results are a symptom of this inseparable feature axis, not merely a byproduct of restricted model capacity.

All scoring heads in our main experiments are sub-1\,B-parameter open-source models that fit within $\leq\!8$\,GB of VRAM. 
As mentioned earlier, this is a deliberate methodological choice reflecting realistic deployment constraints. 
More importantly, it isolates the fundamental theoretical question---whether the underlying \emph{likelihood feature axis} is statistically separable in highly constrained registers---from the orthogonal question of how well specific multi-billion parameter checkpoints calibrate to it. Appendix~\ref{sec:app_modsub} substantiates this substitution through a complete re-evaluation using the original frontier scoring models (e.g., Falcon-7B, GPT-J-6B) on NVIDIA H100 GPUs, confirming that the Perplexity Trap persists and the catastrophic false positive rates are invariant to model scale.

We exclude proprietary API detectors (\href{https://gptzero.me/}{GPTZero}, \href{https://copyleaks.com/}{Copyleaks}) from this evaluation. Their scoring functions are unpublished, their thresholds change without notice, and \cite{barrot} have already documented their catastrophic degradation on templated text. Including them would not add information about the underlying feature axis, which is the object of our study.

\paragraph{Note on Scoring Model Substitution}

The main body of the paper restricts every reported detector number to a realistic deployment infrastructure (single consumer-grade GPU with $\le\!8$\,GB VRAM), because that is a good example of the hardware which non-tech entities like patent offices, IP law firms, and SME R\&D departments would realistically possess. The reference implementations of the detectors benchmarked in this paper, however, utilize substantially larger scoring models: GPT-J-6B for DetectGPT and Fast-DetectGPT, and a dual-model Falcon-7B stack for Binoculars. Because a single consumer GPU cannot host these models in FP16, we substituted the GPT-2 family (124M and 355M) for the main experiments. This substitution fulfills the memory constraints, satisfies Binoculars' joint-tokenization invariant, and avoids advantaging one method over another. 

To ensure this substitution does not invalidate our conclusions regarding the Perplexity Trap, we conducted a comprehensive ablation on an NVIDIA H100 80\,GB GPU, re-evaluating the corpus using the originally published multi-billion parameter scoring heads. 

\textbf{What this appendix is not.} This is a register-matched head restoration on the paper's EPO H04 corpus, not a reproduction of each detector's original paper on its native benchmarks. The question answered here is the following: \emph{does restoring the published Hugging Face stacks on datacenter hardware yield a working solution on patent claims and abstracts?}

\paragraph{1. Per-detector Spearman Rank Correlation.}
We first computed the per-document Spearman rank correlation $\rho$ between the substitute-head scores and the original-head scores (Table~\ref{tab:spearman_substitute_vs_original}). The correlation is strong for Binoculars ($\rho=0.84$ on claims) and Fast-DetectGPT ($\rho=0.78$ on claims), indicating that the smaller models preserve the inferential ordering of the likelihood axis. While DetectGPT exhibits a weaker correlation ($\rho=0.39$), the overall rank-based statistics remain robust. Distributional inseparability, once established, is largely invariant to monotonic transformations of the score axis induced by larger models.

\begin{table}[ht]
	\centering
	\small
	\caption{Per-detector Spearman rank correlation $\rho$ between substitute-head scores (deployment frontier, $\le\!8$\,GB VRAM) and originally-published-head scores (NVIDIA H100 80\,GB) on the same EPO H04 corpus. Both score sets are computed on the identical $1{,}000$ documents (abstracts) or $11{,}613$ claims; the join key is (publication\_number, claim\_number); $n$ is the number of overlapping (document, score) pairs.}
	\label{tab:spearman_substitute_vs_original}
	\begin{tabularx}{\textwidth}{@{} l l c >{\raggedright\arraybackslash}X >{\raggedright\arraybackslash}X c @{}}
		\toprule
		\textbf{Detector} & \textbf{Granularity} & \textbf{$n$} & \textbf{Substitute head} & \textbf{Original head} & \textbf{Spearman $\rho$} \\
		\midrule
		Binoculars     & abstracts & 1000  & GPT-2 / GPT-2-medium    & Falcon-7B / Falcon-7B-Instruct & 0.786 \\
		Binoculars     & claims    & 11613 & GPT-2 / GPT-2-medium    & Falcon-7B / Falcon-7B-Instruct & 0.836 \\
		Fast-DetectGPT & abstracts & 1000  & GPT-2-medium            & GPT-J-6B                       & 0.669 \\
		Fast-DetectGPT & claims    & 11613 & GPT-2-medium            & GPT-J-6B                       & 0.780 \\
		DetectGPT      & abstracts & 1000  & GPT-2-medium / T5-small & GPT-J-6B / T5-large            & 0.115 \\
		DetectGPT      & claims    & 11613 & GPT-2-medium / T5-small & GPT-J-6B / T5-large            & 0.389 \\
		\bottomrule
	\end{tabularx}
\end{table}

\paragraph{2. Head-restored Re-evaluation at Published Thresholds.}
More critically, we demonstrate that the detection failure is not an artifact of scoring head capacity. When the original Falcon-7B and GPT-J-6B scoring stacks are deployed on the H100 GPU to evaluate the EPO patent register at their published default thresholds, the catastrophic failure modes persist. 


\begin{table}[ht]
  \centering
  \small
  \begin{tabular}{l l r r r r r r r r}
    \toprule
    \textbf{Detector} & \textbf{Granularity} & \textbf{N} & \textbf{TP} & \textbf{FP} & \textbf{FN} & \textbf{TN} & \textbf{Acc.} & \textbf{FPR} & \textbf{AUROC} \\
    \midrule
    Binoculars & abstracts & 1000 & 0 & 0 & 500 & 500 & 0.500 & 0.000 & 0.623 \\
    Binoculars & claims & 11613 & 0 & 0 & 5000 & 6613 & 0.569 & 0.000 & 0.701 \\
    Fast-DetectGPT & abstracts & 1000 & 28 & 74 & 472 & 426 & 0.454 & 0.148 & 0.542 \\
    Fast-DetectGPT & claims & 11613 & 972 & 3376 & 4028 & 3237 & 0.362 & 0.511 & 0.743 \\
    DetectGPT & abstracts & 1000 & 453 & 444 & 47 & 56 & 0.509 & 0.888 & 0.626 \\
    DetectGPT & claims & 11613 & 4327 & 5147 & 673 & 1466 & 0.499 & 0.778 & 0.591 \\
    \bottomrule
  \end{tabular}
  \caption{Confusion matrices, accuracy, false-positive rate, and threshold-free AUROC of the three benchmarked detectors on the EPO H04 corpus, evaluated on a single NVIDIA H100 80\,GB HBM3 GPU with the originally published scoring heads (Falcon-7B + Falcon-7B-instruct for Binoculars; GPT-J-6B for Fast-DetectGPT; GPT-J-6B + T5-large mask model for DetectGPT)}
  \label{tab:nml-a100-confmat}
\end{table}

Table~\ref{tab:nml-a100-confmat} lists TP/FP/FN/TN, accuracy, \ac{fpr}, and threshold-free \ac{auroc} for each detector at abstract and claim level granularity under the originally published heads.

\begin{figure}[ht]
	\centering
	\footnotesize
	\begin{subfigure}{0.31\linewidth}
		\centering
		\includegraphics[width=\linewidth]{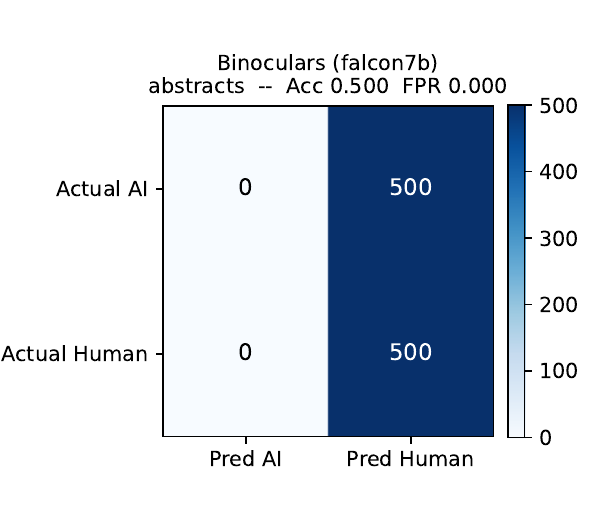}
		\caption{Binoculars, abstracts}
	\end{subfigure}\hfill
	\begin{subfigure}{0.31\linewidth}
		\centering
		\includegraphics[width=\linewidth]{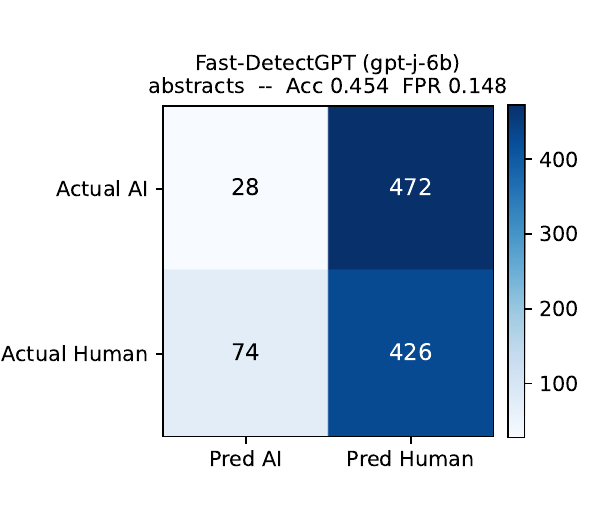}
		\caption{Fast-DetectGPT, abstracts}
	\end{subfigure}\hfill
	\begin{subfigure}{0.31\linewidth}
		\centering
		\includegraphics[width=\linewidth]{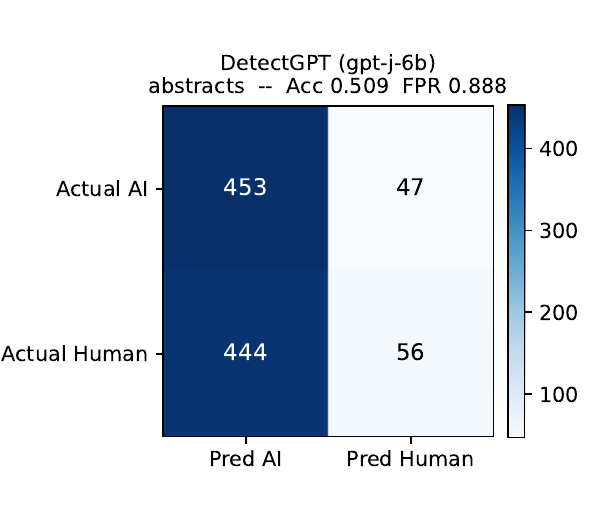}
		\caption{DetectGPT, abstracts}
	\end{subfigure}
	
	\medskip
	
	\begin{subfigure}{0.31\linewidth}
		\centering
		\includegraphics[width=\linewidth]{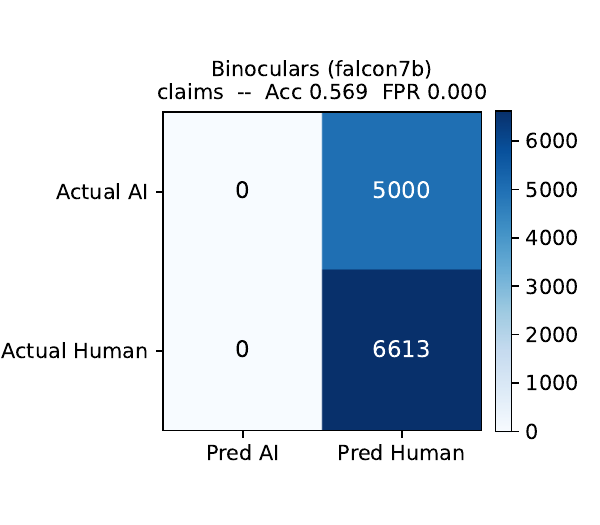}
		\caption{Binoculars, claims}
	\end{subfigure}\hfill
	\begin{subfigure}{0.31\linewidth}
		\centering
		\includegraphics[width=\linewidth]{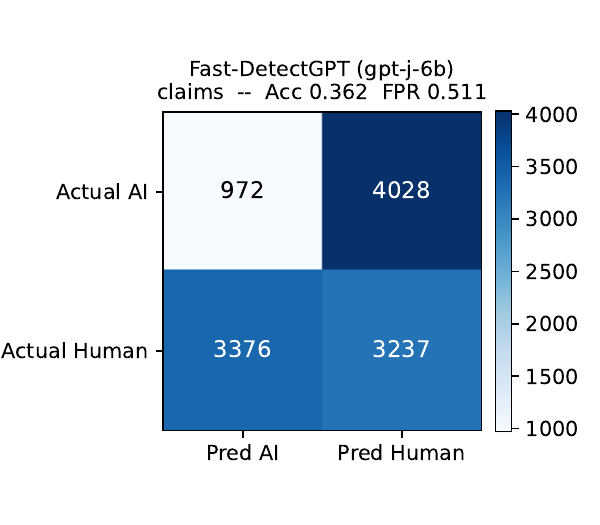}
		\caption{Fast-DetectGPT, claims}
	\end{subfigure}\hfill
	\begin{subfigure}{0.31\linewidth}
		\centering
		\includegraphics[width=\linewidth]{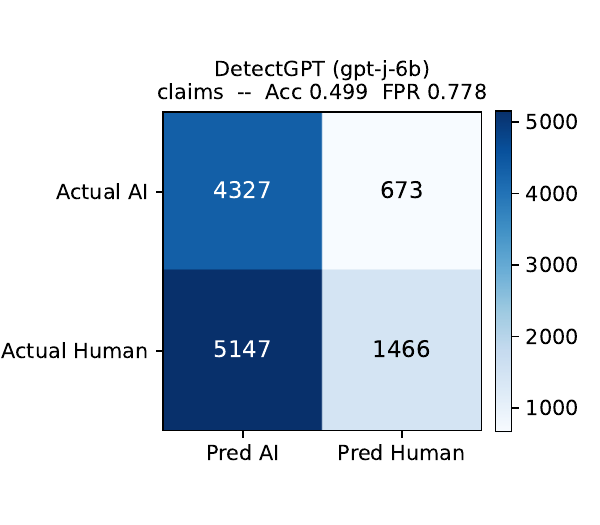}
		\caption{DetectGPT, claims}
	\end{subfigure}
	\caption{Confusion-matrix heatmaps for the six runs on the EPO H04 corpus with originally published scoring heads.}
	\label{fig:nml-confmats}
\end{figure}

As shown in Figure~\ref{fig:nml-confmats}, each model exhibits distinct pathological classification behaviors. Binoculars operates at an impractically conservative threshold; while it successfully protects human authors by producing zero false positives, it entirely fails to identify any AI-generated text (TP = 0), effectively acting as a blind classifier defaulting to a "human" label. Conversely, DetectGPT suffers from catastrophic over-sensitivity, indiscriminately flagging text to produce unacceptably high false positive rates (0.888 for abstracts, 0.778 for claims). Fast-DetectGPT demonstrates erratic calibration; despite achieving the highest underlying class separation on claims (AUROC = 0.743), its practical thresholding yields a degrading overall accuracy of 0.362 and a 0.511 FPR. 

\paragraph{3. Threshold Recalibration.}
To definitively decouple the model capacity from the threshold selection, we re-fit the decision thresholds on a held-out 80/20 stratified split (Table~\ref{tab:recalibration_sweep}). Even at the Youden-optimal threshold (maximizing TPR -- FPR), or at the minimal-FPR threshold satisfying a $0.5$ recall floor, no combination of detector, granularity, and scoring head delivers both a non-trivial recall and a viable ($<5\%$) false positive rate. 

\begin{table}[ht]
	\centering
	\small
	\caption{Best-case operating points after re-fitting the decision threshold on a held-out 80/20 stratified split of the EPO H04 corpus, for both deployment-frontier substitute-head stacks and originally-published-head stacks. Even at the Youden-optimal threshold, or at the minimal-FPR threshold that satisfies a recall floor of $0.5$, no row simultaneously delivers a non-trivial recall and a sub-$5\%$ FPR.}
	\label{tab:recalibration_sweep}
	\begin{tabularx}{\textwidth}{@{} l l l c >{\centering\arraybackslash}X >{\centering\arraybackslash}X >{\centering\arraybackslash}X >{\centering\arraybackslash}X @{}}
		\toprule
		\textbf{Det.} & \textbf{Gran.} & \textbf{Head} & \textbf{$n$} & \textbf{AUROC} & \textbf{Youden FPR} & \textbf{Youden Rec.} & \textbf{FPR @ Rec.$\ge 0.5$} \\
		\midrule
		Binoculars     & abstracts & substitute & 1000  & 0.535 & 0.200 & 0.306 & 0.498 \\
		Binoculars     & claims    & substitute & 11613 & 0.647 & 0.276 & 0.548 & 0.227 \\
		Binoculars     & abstracts & original   & 1000  & 0.623 & 0.400 & 0.600 & 0.320 \\
		Binoculars     & claims    & original   & 11613 & 0.701 & 0.400 & 0.730 & 0.248 \\
		Fast-DetectGPT & abstracts & substitute & 1000  & 0.387 & 0.060 & 0.086 & 0.728 \\
		Fast-DetectGPT & claims    & substitute & 11613 & 0.268 & 0.031 & 0.058 & 0.891 \\
		Fast-DetectGPT & abstracts & original   & 1000  & 0.458 & 0.814 & 0.896 & 0.588 \\
		Fast-DetectGPT & claims    & original   & 11613 & 0.257 & 0.022 & 0.104 & 0.913 \\
		DetectGPT      & abstracts & substitute & 1000  & 0.808 & 0.320 & 0.788 & 0.144 \\
		DetectGPT      & claims    & substitute & 11613 & 0.542 & 0.367 & 0.471 & 0.398 \\
		DetectGPT      & abstracts & original   & 1000  & 0.626 & 0.228 & 0.360 & 0.316 \\
		DetectGPT      & claims    & original   & 11613 & 0.591 & 0.419 & 0.570 & 0.358 \\
		\bottomrule
	\end{tabularx}
\end{table}

\textbf{Implication.} These results confirm that the Perplexity Trap is a fundamental failure of the likelihood axis under statutory constraint, not merely a symptom of restricted scoring model capacity or sub-optimal thresholding. Substitution of smaller models on the deployment frontier preserves the structural inseparability of the domain.
\section{Perplexity and Burstiness}
\label{sec:app_perp_bursti}

In the computational analysis of generated text, perplexity and burstiness serve as primary statistical discriminators between human and non-human (e.g., AI) authorship. 
Perplexity quantifies the predictive uncertainty of an autoregressive language model evaluated on a given sequence, formally defined as the exponentiated average negative log-likelihood of the token sequence under the model's learned distribution, 
$\text{PPL}(X) = \exp(-\frac{1}{N} \sum_{i=1}^N \log p_\theta(x_i|x_{<i}))$. 
Consequently, text generated by \acp{llm} typically exhibits anomalously low perplexity when scored by structurally similar models, as the sequence strictly adheres to the high-probability manifold of the training distribution. 
Burstiness, in this context, characterizes the variance of these linguistic features---specifically the fluctuation of sentence-level perplexity or structural complexity---across a document's span. 
While machine-generated prose tends to optimize for uniform token predictability, resulting in a consistent, flat structural cadence (low burstiness), human authorship natively demonstrates high structural heterogeneity, manifesting as unpredictable ``bursts'' of complex, low-probability syntax interspersed with highly probable, conventional phrasing.
\section{Replication on Additional Registers}
\label{sec:app_extended}

To establish that perplexity trap, as written in Section~\ref{sec:problem} is a property of the register-constraint condition (C1)--(C3) rather than that of the specific EPO H04 telecoms corpus, we replicated the main experimental pipeline on additional registers spanning different technical domains. 

The AI side of every additional corpus was generated by Qwen2.5-3B-Instruct (Alibaba) under the same five prompting strategies as Cat~A--E in Appendix~\ref{sec:app_prompts}; usage of a different LLM serves as an additional cross-LLM robustness check.

\subsection{Replication on Non-Telecom Patents (A61K + C07D + F03D)}
\label{sec:app_extended_patents_mixed}

We replicated the main detector benchmark on a fresh corpus of 500 EPO B1 patents drawn from three vocabulary-orthogonal IPC classes (A61K medical preparations, C07D heterocyclic chemistry, F03D wind motors), all granted before 2020-01-01. The AI side was generated with Qwen2.5-3B-Instruct under the same five prompting strategies (Cat~A--E) used for the main corpus. Mean detector FPR is 84.6\%, the human/AI perplexity overlap is 0.78 (Weitzman coefficient), and the off-axis linguistic-complexity classifier reaches 89.8\% accuracy at 16.0\% FPR --- consistent with the H04 telecoms corpus and confirming that the Perplexity Trap is a property of the patent register rather than of any particular technical vocabulary.

\begin{table}[ht]
    \centering
    \caption{Detector accuracy and false-positive rate on the replication on non-telecom patents (a61k + c07d + f03d) corpus (Qwen-generated AI side).}
    \begin{tabular}{lcc}
        \toprule
        \textbf{Detector} & \textbf{Acc.} & \textbf{FPR} \\
        \midrule
        Binoculars & 52.0\% & 93.4\% \\
        Fast-DetectGPT & 66.2\% & 65.0\% \\
        DetectGPT & 52.2\% & 95.4\% \\
        \bottomrule
    \end{tabular}
\end{table}

\section{Anticipated Reviewer Concern: Why Not Use DivScore?}
\label{sec:app_divscore}

\textbf{Concern} 

A reviewer may reasonably ask why the three detectors evaluated in the main paper do not include DivScore \cite{divscore}, the recent EMNLP-2025 detector that is specifically engineered for specialised legal and medical text. 
Section~\ref{sec:relatedwork} cites DivScore as motivating prior art but does not benchmark it, which could potentially be misread as an evaluation gap. 
We address it here directly by running DivScore on the same EPO\,+\,Cat~A--E\,+\,Cat~F corpus.

\textbf{DivScore Overview}

DivScore \cite{divscore} computes a per-document score
\begin{equation*}
  \text{DivScore}(x) \;=\; \frac{H_{M^*}(x)}{CE_{M^*}(x)},
\end{equation*}
where $H_{M^*}(x)$ is the mean per-position predictive entropy of a domain-adapted student LM $M^*$ and $CE_{M^*}(x)$ is the mean per-token negative log-likelihood of $x$ under $M^*$. 
The decision rule is $\text{DivScore}(x) > \tau \Rightarrow$ AI. 
The published implementation uses Mistral-7B (or other 7B-class LMs) as $M^*$, fine-tuned via knowledge distillation on a domain-specific unlabeled corpus. 
By construction, DivScore is intended to escape the ``distributional collapse'' that its authors identify as the failure mode of likelihood-based detectors in specialised domains.

\paragraph{Deployment Setup}
We maintain the strict compute envelope established previously (a single consumer-grade GPU with $\le\!8$\,GB VRAM, prohibiting external API calls). Because the 7-billion-parameter Mistral model utilized in the original DivScore framework \citep{divscore} exceeds this memory budget, we substitute the largest open-weight causal language model that natively fits: \texttt{EleutherAI/pythia-2.8b-deduped} in half-precision (FP16). We evaluate two configurations of $M^*$, both strictly within the deployment frontier:

\begin{enumerate}
	\item \textbf{DivScore-base}: $M^*$ acts as the un-adapted Pythia-2.8B-deduped checkpoint. This isolates the baseline performance, corresponding to the ``without domain adaptation'' ablation reported by \cite{divscore}.
	\item \textbf{DivScore-adapted}: $M^*$ is the Pythia-2.8B-deduped model LoRA-fine-tuned on $2,500$ held-out human-authored EPO claims. Because we rely on a genuine human corpus rather than synthetic LLM outputs, this configuration serves as the deployment-frontier analogue to DivScore's ``human text finetuning'' baseline, allowing us to test the impact of register alignment on the entropy scoring.
\end{enumerate}

For each text $x$ in EPO, Cat~A--E (claims and abstracts; $11613$ claims and $1000$ abstracts in total) and Cat~F ($923$ Qwen2.5-3B-Instruct claims), we compute DivScore from a single forward pass through $M^*$. 
We report AUROC, accuracy, FPR, and FNR at the Youden-$J$-optimal threshold --- i.e.\ the most favourable evaluation possible for the detector, since DivScore is allowed to choose the best threshold per granularity \emph{after} seeing the test labels.
By giving DivScore the most unfair advantage possible, any failure it exhibits under these conditions is proven to be a fundamental flaw in the metric itself, not just a result of a poorly chosen threshold.
The three detectors of the main paper are evaluated at their published thresholds and receive no such concession.

\textbf{Result} 

\begin{figure}[ht]
\centering
\begin{subfigure}{0.48\textwidth}
\includegraphics[width=\linewidth]{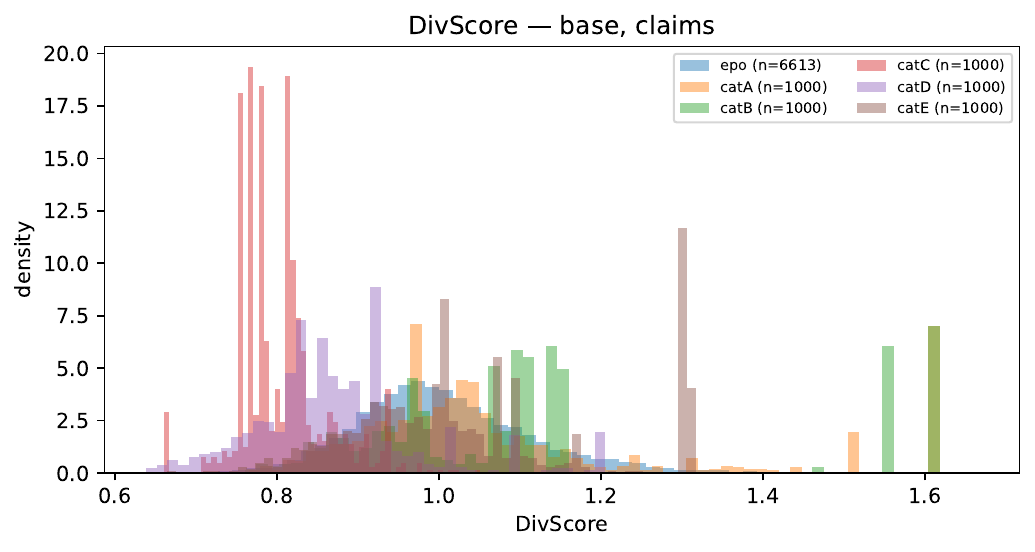}
\caption{DivScore (base $M^*$), claims}
\label{fig:divscore-dist-base-claims}
\end{subfigure}
\hfill
\begin{subfigure}{0.48\textwidth}
\includegraphics[width=\linewidth]{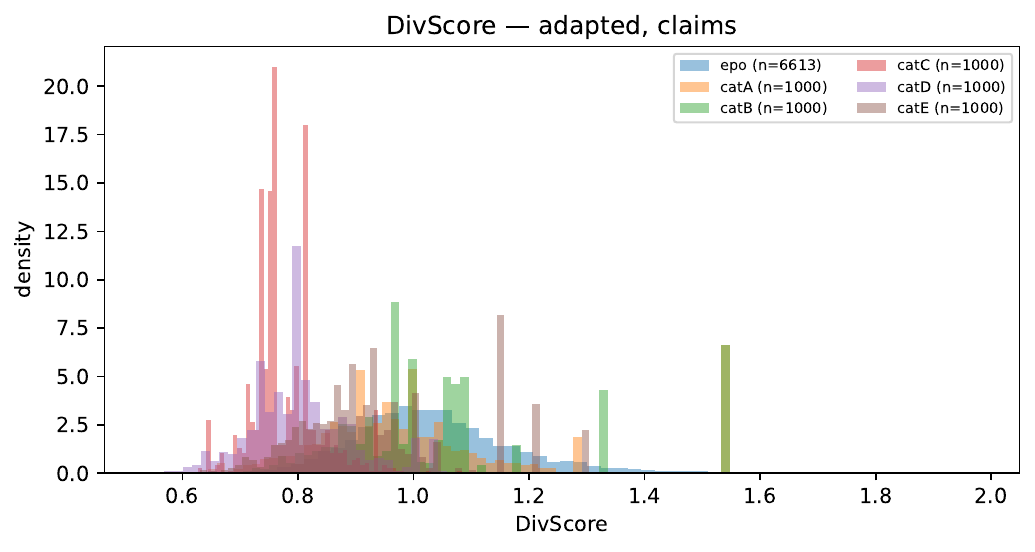}
\caption{DivScore (LoRA-adapted $M^*$), claims}
\label{fig:divscore-dist-adapted-claims}
\end{subfigure}
\caption{DivScore distributions on EPO and Cat~A--E. Even with a domain-adapted $M^*$, the EPO and AI distributions remain largely overlapping; in the base configuration the AI categories straddle the EPO mean (Cat~A, B, E above; Cat~C, D below), so no single threshold $\tau$ on the DivScore axis can separate human from AI. This is the empirical signature of the Perplexity Trap (Section~\ref{sec:perplexity-trap}) operating on the DivScore axis.}
\label{fig:divscore-distributions}
\end{figure}

\begin{figure}[ht]
\centering
\includegraphics[width=0.85\columnwidth]{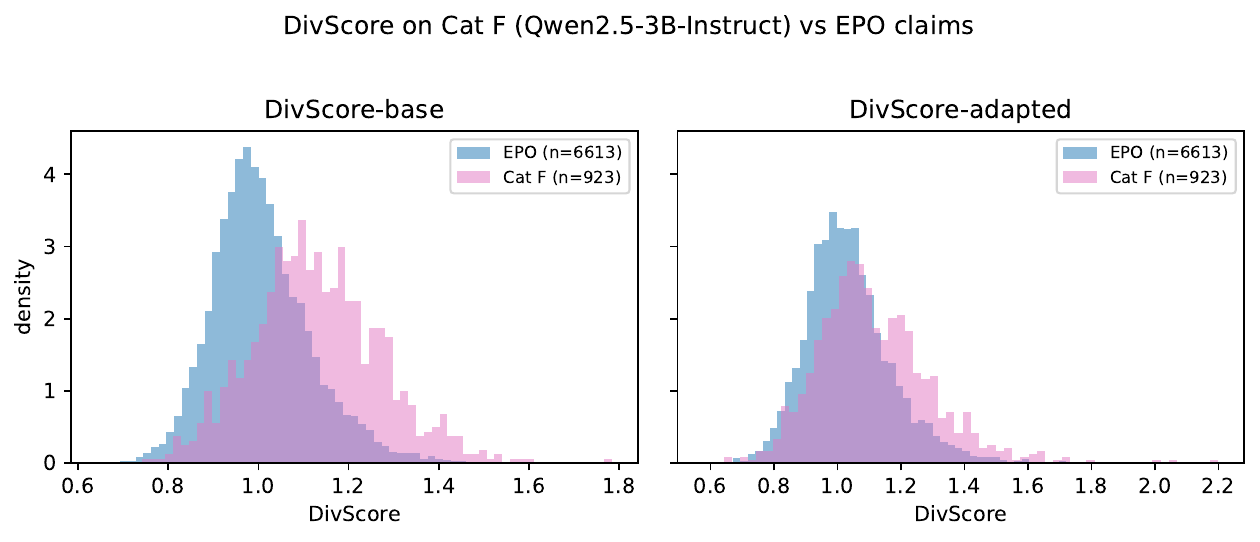}
\caption{DivScore distributions on Cat~F (Qwen2.5-3B-Instruct claims) versus EPO claims, in both $M^*$ configurations. Domain adaptation \emph{worsens} the separation: overlap rises from $0.585$ (base) to $0.780$ (adapted) and $|d|$ falls from $0.96$ to $0.49$. The adapted scoring head has internalised the same surface statistics that Qwen2.5-3B-Instruct itself produces, so the human and AI distributions become more, not less, indistinguishable. This is the same pattern as the raw-perplexity result of Appendix~\ref{sec:app_perp_bursti}, now reproduced on the DivScore axis.}
\label{fig:divscore-catf}
\end{figure}

\begin{figure}[ht]
\centering
\includegraphics[width=0.95\columnwidth]{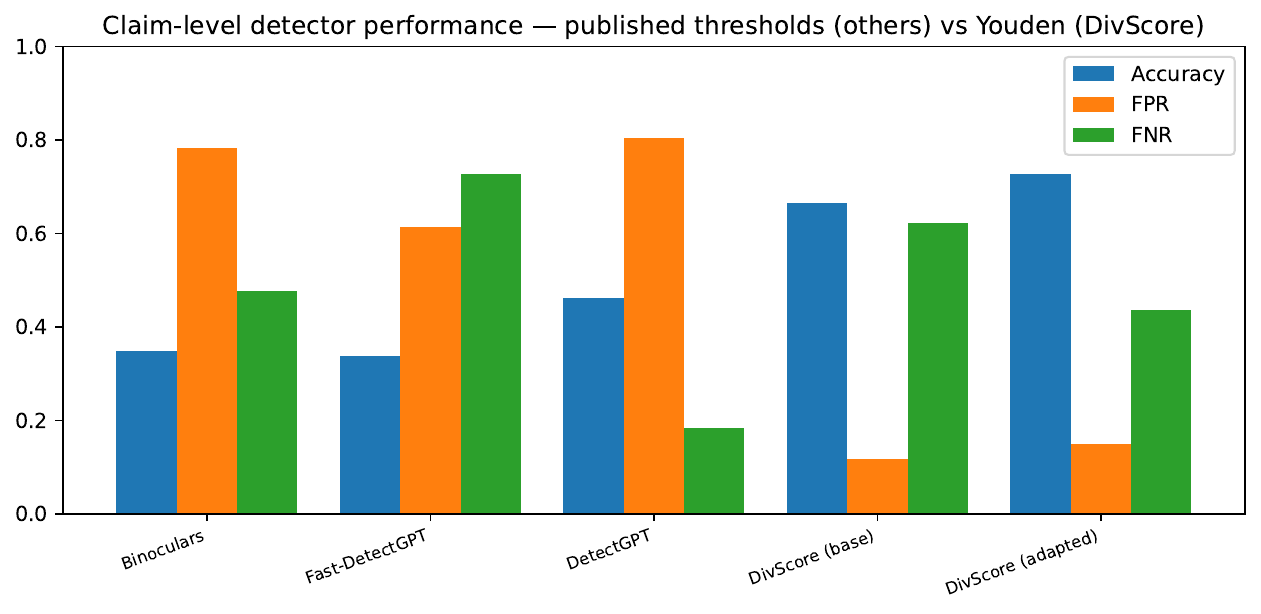}
\caption{Claim-level detector performance: DivScore at the Youden-optimal threshold versus DetectGPT, Fast-DetectGPT and Binoculars at their published thresholds. DivScore receives the most favourable possible evaluation (its threshold is chosen post-hoc on the test set); the other detectors do not. Even under this concession, neither DivScore configuration produces a deployment-acceptable joint $(\text{FPR},\text{FNR})$ pair.}
\label{fig:divscore-detector-comparison}
\end{figure}

Table~\ref{tab:divscore-headline} reports DivScore against the three detectors of Table~\ref{tab:detect_confmat}. Table~\ref{tab:divscore-overlap} reports the per-category distributional overlap on the DivScore axis. Figures~\ref{fig:divscore-distributions}--\ref{fig:divscore-detector-comparison} visualise the result.

\begin{table}[ht]
\centering
\caption{Detector performance at the deployment frontier. The first three detectors are evaluated at their published decision thresholds (reproduced from Table~\ref{tab:detect_confmat}). The two DivScore configurations are evaluated at the Youden-$J$-optimal threshold per granularity, i.e.\ the post-hoc best-of-test-set threshold; this is a strictly more favourable evaluation than the other detectors receive. Lower FPR and FNR are better.}
\label{tab:divscore-headline}
\begin{tabularx}{\textwidth}{l l X X X X}
\toprule
\textbf{Detector} & \textbf{Granularity} & \textbf{Accuracy (\%)} & \textbf{FPR (\%)} & \textbf{FNR (\%)} & \textbf{AUROC} \\ \midrule
Binoculars                                       & Abstracts & 48.6 & 54.2 & 48.6 & ---   \\
Binoculars                                       & Claims    & 34.8 & 78.3 & 47.7 & ---   \\
Fast-DetectGPT                                   & Abstracts & 47.4 & 31.2 & 74.0 & ---   \\
Fast-DetectGPT                                   & Claims    & 33.8 & 61.3 & 72.6 & ---   \\
DetectGPT                                        & Abstracts & 61.3 & 75.4 &  2.0 & ---   \\
DetectGPT                                        & Claims    & 46.2 & 80.5 & 18.4 & ---   \\ \midrule
DivScore (base, $M^*$ = Pythia-2.8B)             & Abstracts & 55.7 & 62.2 & 26.4 & 0.523 \\
DivScore (base, $M^*$ = Pythia-2.8B)             & Claims    & 66.6 & 11.6 & 62.3 & 0.604 \\
DivScore (adapted, LoRA on $2500$ EPO claims)    & Abstracts & 64.1 & 47.0 & 24.8 & 0.650 \\
DivScore (adapted, LoRA on $2500$ EPO claims)    & Claims    & 72.6 & 15.1 & 43.6 & 0.747 \\
\bottomrule
\end{tabularx}
\end{table}

\begin{table}[ht]
\centering
\caption{DivScore distributional overlap (Weitzman) and Cohen's $|d|$ between EPO claims and each AI category, in both DivScore configurations. Higher overlap $\Rightarrow$ the human and AI distributions are less separable on the DivScore axis.}
\label{tab:divscore-overlap}
\begin{tabularx}{\textwidth}{l X X X X X X}
\toprule
\textbf{AI cat.} & \textbf{Ovl.\ base (claims)} & \textbf{$|d|$ base (claims)} & \textbf{Ovl.\ adapted (claims)} & \textbf{$|d|$ adapted (claims)} & \textbf{Ovl.\ base (abstracts)} & \textbf{Ovl.\ adapted (abstracts)} \\ \midrule
Cat A & 0.760 & 0.50 & 0.661 & 0.11 & 0.560 & 0.498 \\
Cat B & 0.496 & 0.67 & 0.540 & 0.10 & 0.502 & 0.402 \\
Cat C & 0.195 & 2.33 & 0.161 & 2.41 & 0.440 & 0.356 \\
Cat D & 0.374 & 1.21 & 0.261 & 1.93 & 0.556 & 0.420 \\
Cat E & 0.675 & 0.22 & 0.463 & 0.70 & 0.416 & 0.502 \\ \midrule
\textbf{Cat F (Qwen)} & \textbf{0.585} & \textbf{0.96} & \textbf{0.780} & \textbf{0.49} & --- & --- \\
\bottomrule
\end{tabularx}
\end{table}

\begin{figure}[ht]
\centering
\includegraphics[width=0.95\columnwidth]{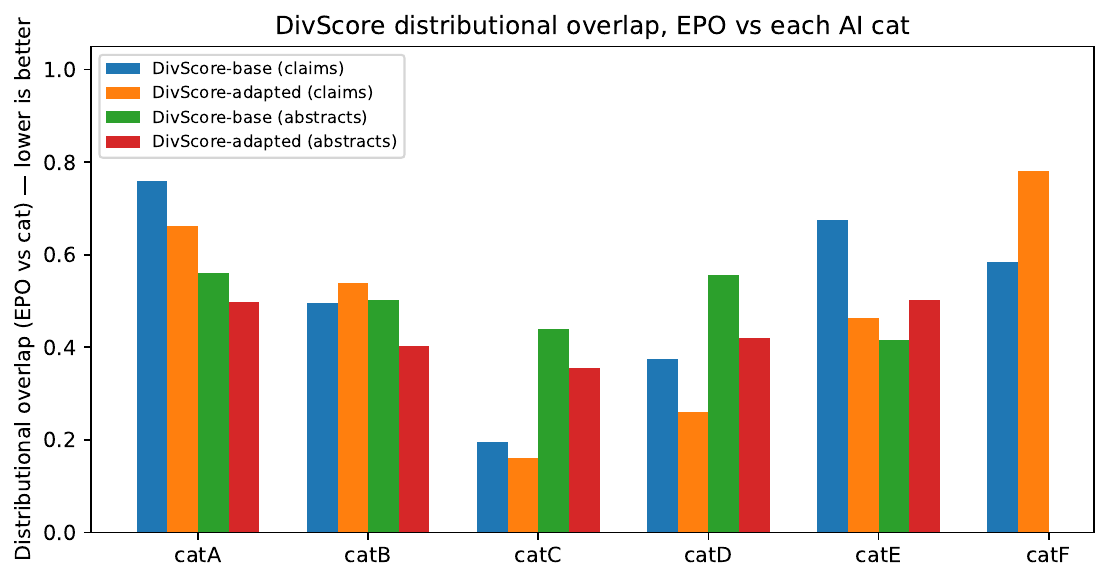}
\caption{Per-category distributional overlap on the DivScore axis, all four configurations. Domain adaptation reduces overlap on the templated Claude categories (most strongly on Cat~D and Cat~E) but \emph{increases} overlap on Cat~F (Qwen2.5-3B-Instruct), the only AI subcorpus produced by a contemporary instruction-tuned LLM in the dataset.}
\label{fig:divscore-overlap-per-cat}
\end{figure}

\textbf{Interpretation of the Result.} Three observations follow from Tables~\ref{tab:divscore-headline}--\ref{tab:divscore-overlap} and Figures~\ref{fig:divscore-distributions}--\ref{fig:divscore-overlap-per-cat}. \textit{(i)} Even when DivScore is allowed to choose the best possible threshold post-hoc on the test set --- a concession that none of the other detectors receive --- it does not produce an operationally usable joint $(\text{FPR},\text{FNR})$ at the claim level: the best configuration (DivScore-adapted) reaches $(15.1\%,\;43.6\%)$, i.e.\ the detector flags roughly one in seven human-authored claims as AI while still letting more than two in five AI claims through. This is an improvement over DetectGPT's $(80.5\%,\;18.4\%)$ at the published threshold, but neither pair is acceptable at examiner workstations where the cost of a false-positive inventorship-attribution error is asymmetric. \textit{(ii)} The per-category DivScore distributions on the patent corpus exhibit the same straddling-the-EPO-mean pattern that raw perplexity does in Section~\ref{sec:perplexity-trap}. In the base configuration on claims, the EPO median is $0.99$ and the AI medians are $1.03$ (Cat~A), $1.09$ (Cat~B), $0.80$ (Cat~C), $0.86$ (Cat~D), $1.01$ (Cat~E) and $1.12$ (Cat~F): four categories sit \emph{above} EPO and two sit \emph{below}. The orientation ``DivScore $<\tau\Rightarrow$ AI'' required by the published decision rule is therefore violated by 66.2\% of the AI population in the corpus (Cat~A, B, E, F), regardless of the choice of $\tau$. Adaptation aligns the orientation (all AI medians shift below EPO) but at the cost described next. The adapted DivScore axis, in other words, has been pulled \emph{closer} to the contemporary AI distribution, exactly as the Perplexity Trap predicts.

\subsection{LoRA training details (DivScore-adapted)}
\label{app:lora-details}

We adapt \texttt{EleutherAI/pythia-2.8b-deduped} as the DivScore student $M^{*}$ using Hugging Face \texttt{transformers} + \texttt{peft}. The base checkpoint is loaded in FP16 with gradient checkpointing enabled, LoRA adapters are attached to the GPT-NeoX attention projection (\texttt{query\_key\_value}) only, and we train for a single epoch on a \emph{train-split} subsample of at most 2{,}500 EPO claims. The held-out test patents (20\% of EPO publications, split at the publication level with seed 42 to avoid claim-level leakage) are \emph{not} seen during fine-tuning. All hyperparameters are listed in Table~\ref{tab:lora-hparams}; values not listed are \texttt{transformers.TrainingArguments} defaults at the pinned versions reported here (\texttt{transformers}~5.5.4, \texttt{peft}~0.19.1, PyTorch~2.5, CUDA~12.1). At DivScore inference time we load the saved adapter via \texttt{PeftModel.from\_pretrained} (no weight merge) and truncate inputs at 256 tokens, as in the un-adapted baseline.

\begin{table}[h]
	\centering
	\caption{LoRA fine-tuning hyperparameters for DivScore-adapted $M^{*}$ on EPO claims. Values not listed match \texttt{transformers.TrainingArguments} defaults.}
	\label{tab:lora-hparams}
	\small
	\begin{tabular}{ll}
		\toprule
		Hyperparameter & Value \\
		\midrule
		Base model & \texttt{EleutherAI/pythia-2.8b-deduped} (FP16) \\
		LoRA rank $r$ & 16 \\
		LoRA $\alpha$ & 32 \\
		LoRA dropout & 0.05 \\
		LoRA target modules & \texttt{query\_key\_value} \\
		LoRA bias & \texttt{none} \\
		Task type & \texttt{CAUSAL\_LM} \\
		Optimizer & AdamW (\texttt{adamw\_torch}, HF defaults) \\
		Weight decay & 0.0 \\
		Peak learning rate & $2\times 10^{-4}$ \\
		LR schedule & linear decay, 0 warmup \\
		Per-device batch size & 2 \\
		Gradient accumulation & 4 \\
		Effective batch size & 8 \\
		Epochs & 1 \\
		Max sequence length & 128 tokens (pad + truncate) \\
		Gradient clipping & 1.0 (default) \\
		Precision & FP16 \\
		Random seed & 42 \\
		Saved artifact & PEFT adapter + tokenizer (no merge) \\
		\bottomrule
	\end{tabular}
\end{table}

\section{Data Generation Prompts}
\label{sec:app_prompts}

The AI-generated patent datasets were generated using Claude Opus 4.6. Batch processing and prompt orchestration were executed via the Cursor IDE (Version 3.0.16) in April 2026. For the categories mentioned in Table~\ref{tab:ai-patent-categories}, the following prompts were used.

\subsection*{Category A (Standard Zero-shot AI)}

\textbf{Goal}: Establish the baseline detection rate for standard, unprompted LLM output.

\textbf{Prompt}: ``1. Write a European patent disclosure.\\
2. Based on the title and the problem statement, write a patent abstract.\\
3. Based on the abstract, draft a standard set of 10 patent claims, including at least one independent method claim and one independent system claim. Use standard patent drafting terminology."

\subsection*{Category B (Persona-Shift AI)}

\textbf{Goal}: Mask the neutral AI tone with dense, conversational engineering logic.

\textbf{Prompt}: ``1. Act as a senior telecommunications engineer while drafting a technical disclosure for an EPO patent attorney.\\
2. Based on the title and the problem statement, write a patent abstract.\\
3. Based on the abstract, draft a standard set of 10 patent claims, including at least one independent method claim and one independent system claim. Use standard patent drafting terminology.\\
4. While drafting, prioritize technical accuracy over perfect legal formatting. Describe the components directly (e.g., use 'transceiver assembly' instead of 'communication unit') and ensure the logical flow mimics a system architecture document."

\subsection*{Category C (Rhythm Control AI)}

\textbf{Goal}: Break the statistical rhythm (entropy) that detectors look for by forcing structural constraints.

\textbf{Prompt}: ``1. While drafting a patent disclosure, use the following structural constraints: alternate between extremely short, punchy technical facts and long, complex subordinate clauses.\\
2. Based on the title and the problem statement, write a patent abstract.\\
3. Based on the abstract, draft a standard set of 10 patent claims, including at least one independent method claim and one independent system claim. \\
4. While drafting, avoid standard predictable boilerplate where possible while remaining legally coherent. Inject specific 3GPP-style telecom jargon to increase the technical perplexity of the text."

\subsection*{Category D (Iterative Refinement AI)}

\textbf{Goal}: Simulate a human-in-the-loop workflow by forcing the LLM to process the data twice within the same prompt.

\textbf{Prompt}: ``1. You will execute a two-step drafting process.\\
2. Read the title and the problem statement.\\
3. Generate an unstructured, bulleted list of raw technical features and logical steps. \\
4. Rewrite those raw notes into a formal EPO patent disclosure containing an Abstract and 10 Claims, including at least one independent method claim and one independent system claim. \\
5. Intentionally include two human idiosyncrasies in the final output: use a technical acronym once before defining it, and use slightly asymmetrical phrasing in the dependent claims rather than perfectly uniform templates."

\subsection*{Category E (Non-native AI)}

\textbf{Goal}: Test if detectors unfairly flag non-native technical English by simulating a German or Chinese applicant filing in English.

\textbf{Prompt}: ``1. While drafting a patent disclosure in formal English, act as a non-native English speaking inventor from Germany.\\
2. Based on the title and the problem statement, write a patent abstract. Use highly rigid, grammatically strict, and formal vocabulary. Avoid colloquialisms entirely. Rely heavily on passive voice and highly structured, predictable transitions.\\
3. Based on the abstract, draft a standard set of 10 patent claims, including at least one independent method claim and one independent system claim. \\
4. While drafting, strictly adhere to repetitive phrasing templates for every single claim (e.g., 'The method according to claim 1, further comprising...'). The text should feel technically flawless but stylistically stiff."

\section{Multi-LLM Robustness (Cat~F: Qwen2.5-3B-Instruct)}
\label{sec:app_multillm}

The five AI categories evaluated in Section~\ref{sec:dataset} (Cat~A--E) were all generated by Claude Opus 4.6 (Anthropic). 
One may legitimately ask whether the conclusions of Section~\ref{sec:problem} are an artifact of a single LLM family rather than a fundamental property of the patent domain. 
We therefore generated a sixth category, Cat~F, comprising $100$ patent abstracts and $923$ claims using Qwen2.5-3B-Instruct (Alibaba). Keeping the rest of the pipeline fixed, we re-ran the three primary analyses: raw perplexity overlap, three-detector evaluation, and complexity-feature transfer.

\textbf{Raw perplexity overlap (Cat~F vs.\ EPO)} 

\begin{table}[ht]
	\centering
	\caption{Raw perplexity overlap and effect size between EPO and Cat~F (Qwen2.5-3B-Instruct).}
	\label{tab:catf-overlap}
	\begin{tabularx}{\columnwidth}{l X X X X}
		\toprule
		\textbf{Granularity} & \textbf{Overlap} & \textbf{$|d|$} & \textbf{Hum.\ ppl} & \textbf{AI ppl} \\ \midrule
		Abstracts & 0.48 & 0.99 & 19.95 & 10.89 \\
		Claims    & 0.85 & 0.28 & 17.81 & 14.20 \\
		\bottomrule
	\end{tabularx}
\end{table}

At the claim granularity, the human and Cat~F perplexity distributions overlap by $0.85$ with an effect size of $|d| = 0.28$---comparable to the most highly overlapping Claude category (Cat~A, $0.81$). At the abstract granularity, the overlap is $0.48$ with $|d| = 0.99$ (Table~\ref{tab:catf-overlap}). 
The Perplexity Trap established in Section~\ref{sec:problem} is therefore not a Claude-specific artifact: a completely different LLM family, despite having a fraction of the parameter count of the frontier models used in the primary evaluation, produces text that is statistically inseparable from human-authored EPO claims under the same likelihood axis.

\textbf{Detector Evaluation on Cat~F} 

\begin{table}[ht]
	\centering
	\caption{Three-detector evaluation on EPO+Cat~F.}
	\label{tab:catf-detector}
	\begin{tabularx}{\columnwidth}{l l X X X X}
			\toprule
			\textbf{Detector} & \textbf{Gran.\ } & \textbf{Acc.} & \textbf{Recall} & \textbf{FPR} & \textbf{F1} \\ \midrule
			Binoculars     & Abstracts & 0.700 & 0.980 & 0.580 & 0.766 \\
			Binoculars     & Claims    & 0.484 & 0.850 & 0.782 & 0.581 \\
			Fast-DetectGPT & Abstracts & 0.840 & 0.980 & 0.300 & 0.860 \\
			Fast-DetectGPT & Claims    & 0.487 & 0.605 & 0.599 & 0.498 \\
			DetectGPT      & Abstracts & 0.585 & 1.000 & 0.830 & 0.707 \\
			DetectGPT      & Claims    & 0.511 & 0.934 & 0.797 & 0.617 \\
			\bottomrule
		\end{tabularx}
\end{table}

We re-ran those main detectors from the main body of the paper (Binoculars, DetectGPT,Fast-DetectGPT) on the EPO+CatF mix (Table~\ref{tab:catf-detector}). 
The results closely mirror the collapse observed with the Claude-generated corpus in the main text. While the detectors retain marginal discriminative power on the less-constrained abstracts (with Fast-DetectGPT achieving $0.840$ accuracy), performance degrades catastrophically at the claim granularity. Across all three methods, overall accuracy on patent claims plummets to near-random chance ($0.484$--$0.511$). Crucially, the detectors exhibit a degenerate classification behavior: they achieve seemingly high recall only by indiscriminately flagging text as AI-generated, resulting in severe false positive rates ($0.599$ to $0.797$) on genuine human-authored claims. 
This confirms that the Perplexity Trap fundamentally neutralizes likelihood-based detection, regardless of the underlying generative model's architecture or scale.

\textbf{Complexity-feature Transfer} 

\begin{table}[ht]
	\centering
	\caption{Zero-shot transfer of the complexity-feature LR (trained on EPO + Cat~A--E) to Cat~F.}
	\label{tab:catf-transfer}
	\begin{tabularx}{\columnwidth}{l X X X X X}
		\toprule
		\textbf{$n_{\text{train}}$} & \textbf{$n_{\text{test}}$} & \textbf{Acc.} & \textbf{F1} & \textbf{FPR} & \textbf{Recall} \\ \midrule
		11613 & 2191 & 0.701 & 0.657 & 0.285 & 0.680 \\
		\bottomrule
	\end{tabularx}
\end{table}

Finally, we evaluated whether the complexity-feature classifier introduced in Section~\ref{sec:linguistic-complexity}, which was trained exclusively on the Claude-generated mixture (EPO+Cat~A--E), transfers \emph{zero-shot} to the Qwen-generated Cat~F. As shown in Table~\ref{tab:catf-transfer}, the logistic regression model maintains a viable discriminative signal, achieving an accuracy of $0.701$ and a much more manageable false positive rate of $0.285$. While this represents a degradation compared to strictly in-distribution evaluation---indicating that different LLM families possess distinct stylistic fingerprints---this feature-based approach remains vastly superior to the baseline zero-shot detectors. This confirms that while the Perplexity Trap universally neutralizes likelihood-based metrics across different model architectures, stylistic and structural complexity features retain a meaningful degree of cross-model robustness, even in highly constrained registers.

\textbf{Implication.} The Perplexity Trap, the structural-marker discrimination of M2 and M6, and the complexity-feature lift over a likelihood baseline are all reproduced when the AI text is generated by a different LLM family. None of the qualitative conclusions of the main text depend on having Claude as the AI text source.


\end{document}